\documentclass{article}

\usepackage[preprint]{neurips_2019}

\usepackage[utf8]{inputenc} % allow utf-8 input
\usepackage[T1]{fontenc}    % use 8-bit T1 fonts
\usepackage{hyperref}       % hyperlinks
\usepackage{url}            % simple URL typesetting
\usepackage{booktabs}       % professional-quality tables
\usepackage{amsfonts}       % blackboard math symbols
\usepackage{nicefrac}       % compact symbols for 1/2, etc.
\usepackage{microtype}      % microtypography

\usepackage{caption}
\usepackage{subcaption}
\usepackage{natbib}
\setcitestyle{numbers}

\usepackage{multirow}       % tables

\usepackage[pdftex]{graphicx}
\usepackage[compact]{titlesec}         % you need this package

\titlespacing{\section}{0pt}{0pt}{0pt} % this reduces space between (sub)sections to 0pt, for example
\AtBeginDocument{                      % this will reduce spaces between parts (above and below) of texts within a (sub)section to 0pt, for example - like between an 'eqnarray' and text
  \setlength\abovedisplayskip{0pt}
  \setlength\belowdisplayskip{0pt}
}

\usepackage{float}
\usepackage{subcaption}
\usepackage{caption}
\usepackage{graphicx}
\usepackage{amsmath}
\usepackage{amssymb}
\usepackage{hyperref}
\hypersetup{
    colorlinks=true,
    linkcolor=blue,
    filecolor=magenta,      
    urlcolor=blue,
}

\title{Ensemble Machine Learning Methods for Modeling COVID19 Deaths}

\author{
Rahil Bathwal* \\
California Institute of Technology \\
\texttt{rbathwal@caltech.edu} \\
\And
Pavan Chitta* \\
California Institute of Technology \\
\texttt{pchitta@caltech.edu} \\
\AND
Kushal Tirumala* \\
California Institute of Technology \\
\texttt{ktirumal@caltech.edu} \\
\And
Vignesh Varadarajan*  \\
California Institute of Technology \\
\texttt{vigneshv@caltech.edu} \\
\And
(* equal contribution)
}

\begin{document}

\maketitle

\begin{abstract}
Using a hybrid of machine learning and epidemiological approaches, we propose a novel data-driven approach in predicting US COVID-19 deaths at a county level. The model gives a more complete description of the daily death distribution, outputting quantile-estimates instead of mean deaths, where the model's objective is to minimize the pinball loss on deaths reported by the New York Times coronavirus county dataset. The resulting quantile estimates accurately forecast deaths at an individual-county level for a variable-length forecast period, and the approach generalizes well across different forecast period lengths. We won the Caltech-run modeling competition out of 50+ teams, and our aggregate is competitive with the best COVID-19 modeling systems (on root mean squared error). \footnote{Work was done under the supervision of 
Yaser S. Abu-Mostafa, Professor of Electrical Engineering and Computer Science, California Institute of Technology}
\end{abstract}

\section{Introduction}

With every new pandemic, epidemiologists struggle to come up with models consistent with observed results. With the death toll of the COVID-19 pandemic reaching over 200,000 in US alone, it is crucial that local policymakers have the tools they need to marshal resources and target policies that respond to the severity of the epidemic in their region. Current modeling techniques focus on predicting the course of the epidemic at the state and national level. There is a lack of understanding on how to chart the course of the epidemic on regional areas, like counties in the United States. This poses a huge issue for local governments, since the course of the epidemic can vary widely across these larger areas (especially in large states like California). This makes it difficult to know where to deploy resources, and when to ease social distancing restrictions on a county-basis. 

Current models use either epidemiological techniques or basic machine learning to predict the mean number of deaths. We propose a novel method for predicting deaths from COVID-19 on the county level — we combine state-of-the-art machine learning architectures (LSTM, Gradient Boosted Trees, Random Forests, etc.) and epidemiological techniques (SEIR-QD) to predict quantile estimates for daily deaths, on a variable-length forecast period. This allows us to:
\begin{itemize}
    \item Convey more information about the daily death distribution (quantile forecasts)
    \item Capture the severity of the pandemic on a local scale (county-level)
\end{itemize}

% We found that using the quantile output to output mean deaths gives lower root mean squared error (RMSE) than most other state-of-the-art COVID-19 prediction models. We believe the techniques introduced in this paper can give better predictions in terms of:
% \begin{itemize}
%     \item A more complete description of the death distribution (through quantile forecasts)
%     \item Accuracy (using the quantile forecast to estimate the mean and comparing it to ground-truth daily deaths on a 2-week time period)
% \end{itemize}.

Our goal is to provide policymakers with a model that is useful for predicting the course of the pandemic in their area.

In Section \ref{sec:evaluation} we describe our evaluation metric used to improve and choose models. In Section \ref{sec:datausage} we describe the date processing pipelines we fed into the models. In Section \ref{sec:approach} we describe the architectures for all the machine learning/epidemiological models we use. In Section \ref{sec:results} we describe the results from the various models we tried (as per the evaluation metric in \ref{sec:evaluation}). We then discuss these results in Section \ref{sec:discussion}.
\section{Evaluation}
\label{sec:evaluation}
Most COVID-19 prediction models tend to focus predicting mean mortality rate, in that they output a single number to reflect death count/mortality rate \cite{tuli2020predicting}, \cite{lalmuanawma2020applications}. Our goal is to provide a better description of the death distribution — we approach this by providing an estimate of COVID-19 deaths in the United States with error bars. Specifically, we provide quantile estimates for deaths for the 10th, 20th, 30th, 40th, 50th, 60th, 70th, 80th, and 90th quantile. The metric used to evaluate models was the 9-quantile pinball loss, where for quantile $q \in [0.1, 0.2, 0.3, 0.4, 0.5, 0.6, 0.7, 0.8, 0.9]$, the loss for the quantile $q$ given ground truth $y$ and prediction $\hat{y}$ is:
\vspace{5px}
\[
L_q(y, \hat{y}) = 
\begin{cases}
    (y - \hat{y})q,& y \geq \hat{y}\\
    (\hat{y} - y)(1 - q), & y < \hat{y}
\end{cases}
\]

Pinball loss is one of the common ways to evaluate quantile forecasts \cite{steinwart2011estimating}, since it weights errors depending on the quantile (the error incurred by the prediction $\hat{y}$ will vary as we vary the quantile $\tau$). For example, underpredicting for the $90$th quantile incurs much more error than over predicting; similarly, over predicting for the $10$th quantile incurs more error than underpredicting. You can see in the below figure that for different quantiles, the loss function behaves differently.

\begin{figure}[H]
\centering
\includegraphics[width=0.6\textwidth]{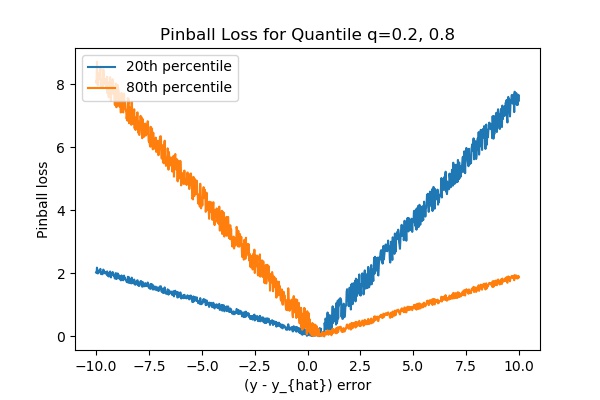}
\caption{A plot showing the pinball loss function for quantiles $q = 0.2$ (blue line) and $q = 0.8$ (orange line). We took the ground truth $y$ as a $1000$ sized vector of random doubles between $0$ and $1$, and $\hat{y}$ as $1000$-sized vector of evenly spaced values from $-10$ and $10$.}
\label{fig:pinballloss}
\end{figure}

Given the quantile estimates for an individual county on a particular day $[\hat{y}_{0.1}, \dots, \hat{y}_{0.9}]$, the pinball loss at a individual county level is:
\vspace{5px}
\[
L_{\text{county}} = \frac{1}{9}\sum_{q \in \{0.1, 0.2, \dots, 0.9\}} L_q(y, \hat{y}_q)
\]

The final evaluation metric is the average of $L_{\text{county}}$ taken over every county, and every day in the forecast period. The ground truth dataset used for evaluation is the New York Times coronavirus dataset \cite{nyt-2020}.

\bigskip

\section{Data Usage and Preprocessing}
\label{sec:datausage}
We describe the data processing and feature engineering done to generate train and test data for all the models used. Labels for daily death values at a county level are obtained from the New York Times Dataset. Initial pre-processing to re-allocate and filter the ground truth data is required due to inconsistent reporting and "data dumps”, wherein counties under-report on certain days and then account for it by accumulating these deaths and over-reporting on a single day. We deal with these "data dumps" by spreading the deaths equally among the days in the period since the last "data dump." \\

This dataset is then combined with the aggregated Berkeley dataset \cite{yu-group_2020}, which captures a wide range of stationary features for each county. As these features are constant for a given county across time, they are much less important than the dynamic features, but certain density features (e.g. population density) proved relevant in differentiating counties. This is finally combined with the Descartes Labs mobility data \cite{descarteslabs_2020}, with missing values imputed in a forward-fill fashion by using the most recent value. The main mobility feature used is a mobility index (m50\_index) which captures a percentage of normal mobility present in a region. \\

The features with the most predictive power are the dynamic features such as cases, mobility and deaths. For several models, these are processed to create derived features with different levels of fixed lag (e.g. 14-days ago deaths, 7-days ago deaths). The amount of lag is constrained by the length of the forecast period and informed by the inherent lag between infection onset and death. For example, for a forecast period of 14 days, the set of lags used (in days) was \{15, 16, 17, 18\}.\\

The focus of the feature engineering is to augment the time-related information available in the data. Such features capture what stage of the pandemic a given county is in. Other information (e.g. day of the week) is also useful in capturing under- and over-reporting trends. Finally, we add categorical features such as the state a county belongs to, as well as the index of the cluster a county belongs to based on the clustering technique described in section \ref{sec:clustering}.

\bigskip

\section{Approach}
\label{sec:approach}
Our modeling approaches can be grouped into two main categories: machine learning models and epidemiological models. While the epidemiological models show promising results in capturing broad trends, they fail to capture any complex patterns in the data; however, the machine learning models are able to capture more granular trends and variations in the data. All of the individual models are combined using a non-linear ensemble described in section \ref{sec:ensemble}.

\subsection{Epidemiological Models}
\label{sec:epidemiologicalmodels}

\subsubsection{Imperial College Model}
\label{sec:imperialcollegemodel}
One of the main models being used is an adaptation of the main Imperial College Model \cite{flaxman2020report}, a semi-mechanistic Bayesian hierarchical model trained to predict daily deaths that is able to incorporate intervention data on dates of lock-down and other closures. The original study was done on 11 European countries; this was adapted to train on US counties instead. The intervention data is used to learn parameters that determine the time-varying reproduction number, and this value is used in combination with an internal infection model to predict the number of true cases and deaths. 

A noteworthy change made to the original model is that multiple parameters are used to learn the uncertainty in death predictions per county, instead of a single shared parameter across all counties. In particular, the top 10 counties by cumulative deaths are trained on individually, separately from the main model used to train on the rest of the counties. In addition, since information on the infection fatality rates (IFR) are not available on a per-county basis, these values are initially set to a common value of the IFR value of the United Kingdom and a multiplicative deviation parameter is learned per county.

\subsubsection{Generalized SEIR Models}
\label{sec:seirmodel}
Another one of our initial models is the SEIR-QD  model \cite{peng2020epidemic}, which is a variant of the generalized SEIR model (discussed in \cite{peng2020epidemic}). The SEIR-QD model is a compartmental model including states for susceptible, exposed, infected, recovered, quarantined, and death with ordinary differential equations governing disease transmission and movement between these states. This variant of SEIR models has been shown to model the COVID-19 epidemic well \cite{yang2020rational}. The model parameters for these ODEs are learnt on a per-county level using a least squares optimization approach to minimize the error in both predicted cases and deaths. For counties in the early stages of the epidemic, a weighted loss function favoring the accuracy of predicted cases is used since the ground truth data on deaths is very noisy. However, for counties with more severe outbreaks, the weighting is shifted to favor accuracy of predicted deaths since the true data on deaths was more reliable. More importantly, an unweighted loss metric yields poor results because the number of reported cases is significantly higher than the number of reported deaths. The prediction uncertainty is generated using the technique based on the negative binomial distribution described in section \ref{sec:negbinomialtrick}.

\subsection{Neural Networks}
\label{sec:nnmodels}

\subsubsection{LSTM}
\label{sec:lstmmodel}
The architecture of the LSTM follows an encoder-decoder approach, taking into account both temporally variable data and fixed variables. Temporal variables used are cases, deaths, a one-hot encoded weekday feature, and fixed variables are population features and hospital bed counts in the county. First, a ConvLSTM with a 5-day convolution window is applied to the data to generate a feature vector. Using this approach over a standard LSTM seemed to help the network handle the noise in the data without having to predict a moving average of deaths, which eliminates some interesting trends that may be beneficial to capture. The output of the ConvLSTM is stacked on top of the vector of fixed features, which is then fed into a dense layer that outputs the final encoding of the death time series.

\textbf{Multiheaded Decoders}
\label{sec:multiheadeddecoders}

Especially early in the epidemic, directly training the encoder-decoder network to predict for the given forecast period from static lagged features (e.g. 14-days-ago cases and deaths), was problematic, as it greatly diminishes the number of training samples we can generate for the network. For example, there are few 14-day periods available in the first 2 months of the epidemic in the United States. To increase the data available to train the encoder segment of the network, multiple decoders are created, each of which is an LSTM that generates an output sequence of $n$, $9 \times 1$ vectors, each representing the quantile predictions on each of the $n$ days in the forecast period. In the $14$-day forecast example, multiple decoder heads (sharing the same encoder backbone) are created for 3, 7, and 14 day time horizons; the 14 day time horizon decoder is used in the final encoder-decoder in the 2-week prediction.

\textbf{LSTM Training Pipeline}
\label{sec:lstmtrainingpipeline}

With the lack of data on larger outbreaks, especially early in the epidemic, a multi-step training pipeline is used to increase the probability of quick convergence on larger datasets for parts of the model, and then fine-tuning on smaller datasets to improve performance on the specific targets.

First the shared encoder is trained along with the shorter timescale decoders for a few epochs, in increasing order of time. Empirically, we found that pre-training the encoder using decoders designed to predict shorter sequences of deaths (in the 14-day forecast example, the 3 and 7 day decoders) achieved optimal performance across stages of the epidemic. Once the encoder is trained with these decoder networks, the final decoder head was attached to the network, and the entire system is trained for a longer period (empirically, about 10 times the number of epochs used to train each smaller timescale works best). Initialization of the network weights by first training for the easier task of predicting shorter sequences improved the final validation performance of the full forecast period network, in addition to the overall stability of training. This effect is especially prevalent in the earlier days of the epidemic when data was sparse (which as mentioned in \cite{yang2020rational} is a difficult part about modeling an epidemic).

This model is then fine-tuned to generate models that predicted deaths on the top 50 counties for the epidemic  (in terms of total deaths) and the less affected counties separately. This improves performance considerably on more affected counties like New York County.

\subsubsection{Standard Neural Network}
\label{sec:standardneuralnetwork}

Motivated by the model ensembling performance described in \ref{sec:ensemble}, one of the individual models used was a standalone neural network. We use a small network with a series of hidden dense layers (three layers of dimensions \{20, 10, 10\} respectively) activated by rectified linear units. Dropout is used for both hidden and input layers. These hyper-parameters are tuned using automated grid searches. Early stopping is also used with a specified tolerance. The model is trained on all counties together to optimize for mean squared error, and a quantile prediction distribution was obtained using the negative binomial method described in section \ref{sec:negbinomialtrick}.

\subsubsection{Quantile Neural Network}
\label{sec:quantilenn}

 The quantile neural network is constructed by training nine independent neural networks, each optimizing for quantile loss of one of the nine quantiles described in section \ref{sec:evaluation} (i.e nine loss functions $L_{0.1}, \dots, L_{0.9}$). The architecture of each of these models is identical to the input data and architecture described in \ref{sec:standardneuralnetwork}. The predictions of these models are finally concatenated to form a quantile forecast.

\subsection{Tree-Based Models}
\label{sec:treebasedmodels}

\subsubsection{Random Forest}
\label{sec:randomforestmodels}

One of the decision tree-based methods that we use is a random forest regressor model. This model performs poorly early on in the pandemic, mainly due to a lack of data (which leads to poor partitioning). The quality of the predictions vastly improves as more data became available. The model is trained on all counties together to minimize MSE. The quantile distribution is obtained by forming an empirical distribution from the predictions of the individual trees. One of the advantages of this model is that it is able to capture weekly reporting trends well. However, it struggles to generate reasonable quantile distributions for some of the top counties due to a lack of data in similar death ranges. This is again an issue of poor partitioning of leaves for very high death values, leading to unstable predictions for counties like New York City. This leads to exploding top quantiles for the highest death counties. Clipped top quantiles are used to combat this.

\subsubsection{Quantile Gradient Boosted Decision Trees}
\label{sec:quantilegbdt}
Another decision tree based method we explored was gradient boosting decision tree models. For each county, $5$ separate GBDT models are trained directly on quantile loss, and the average of those predictions (for every day in the forecast period) is taken. The benefit of this model is that it optimized directly for quantile loss, and predicts different county level trends for each county. The main issue with the predictions is that a lot of predictions end up missing trends across the counties (since this is an individual county level model), and tend to look linear. 

\subsection{Gaussian Processes}
\label{sec:gaussianprocess}
Another model that we explored is a gaussian process regressor (as per \cite{rasmussen2010gaussian}). As with a lot of statistical methods, this method introduces the idea of a marginal likelihood function, where the prior is taken to be a gaussian given by $N(0, K)$ where $K$ is a user defined kernel. This marginal likelihood is maximized via an optimizer (in this case, we used the L-BGFS-B optimizer), where at the end of training we have a predicted mean and variance (which defines a distribution to generate output for the forecast period). Instead of sampling from this distribution directly to generate quantiles, we took the predicted mean, use that as the mean of a negative binomial distribution for the method described in \ref{sec:negbinomialtrick} to generate quantiles. The only significant improvements in the model is given by choice of kernel, specifically a combination of a constant kernel, and a rational quadratic kernel.

\subsection{Ensemble}
\label{sec:ensemble}
Early on, we noticed that different models have unique strengths and weaknesses in their predictions, so we began exploring ensembling techniques that could combine predictions from all of our individual models. Our initial attempts focused on simple methods (linear regression, averaging, and gradient boosted decision tree ensembling) but we found that combining the predictions of these models using a neural network trained directly on pinball loss improved the consistency, stability, and accuracy of the final predictions. We use a shallow network with a series of hidden dense layers activated by rectified linear units (ReLU) with by dropout regularization. Smaller networks improve the reliability of the predictions and consistency of model convergence. Experiments with deeper or more complicated architectural patterns result in over-fitting due to the lack of data. 

The primary input to the aggregation network are the individual model predictions. However, we observed that there were some important features that impact ensemble performance: county characteristics, the number of days into the prediction period, and the day of the week. To help the network identify trends in these features, we provide as additional input one-hot encoded features for: which state the county is in, the county cluster found using the clustering technique in section \ref{sec:clustering}, the number of days into the prediction timeline, and the day of the week.

To generate final predictions, we train each of the individual models daily and formed an aggregation set consisting of the model predictions for models trained for 28 consecutive days. This aggregation set is then used to train the ensemble network, which generates predictions for the forecast period.
\subsubsection{Negative Binomial Technique}
\label{sec:negbinomialtrick}
Many of our individual models are trained to predict mean daily death values. To generate a quantile distribution from these mean predictions, we assumed the daily death values followed a negative binomial distribution with the mean and variance of the daily death random variable $D$ given by:
$$ \mathbb{E}[D] = \mu, \quad \text{Var}[D] = \mu + \frac{\mu^2}{\phi}$$
Here, $\mu$ is the prediction of the model per day, and $\phi$ is a parameter that controls the spread/uncertainty of the distribution. Instead of trying to fit this $\phi$ value, we used a manual heuristic of using the variance of daily deaths in the past $N$ days to estimate the value of $\phi$. Using this parametrization of the distribution, we then sampled from the distribution to generate quantile predictions. This trick is motivated by the sampling method used in \cite{flaxman2020report}.
\subsubsection{Clustering}
\label{sec:clustering}
Initially, we attempted to cluster with dynamic time warping (DTW) as a metric since it has been shown to work well in other time-series tasks, specifically in the audio domain \cite{muda2010voice}, \cite{zhang2014one}. However, clustering with DTW as a metric and K-Means and DBSCAN as clustering algorithms usually assigned all counties to one main cluster and a few small clusters (this behavior persisted across different choices for hyperparameters). We also experimented with using the raw time series data as the featurization for a county, and doing K-Means clustering on that, but that resulted in similar behavior (where most counties get clustered together, and the differences between time-series in individual clusters seemed random).

Instead, we use a more intuitive approach to clustering, motivated by \cite{mahabal2017deep}, wherein we consider changes in time vs. changes in daily death magnitude (this featurizes all time series on a relative scale). Specifically, using the daily death data for each county since March 10, we create a 2D histogram using difference in magnitude, and difference in time. Essentially, we create $dm$ bins (difference in magnitude bins), and $dt$ bins (difference in time bins), given by: \\
\begin{align*}
    dm_{bins} &= [-20, -5],[-5, -2], [-2, -1], [-1, 0], [0, 1], [1, 2], [2, 30], [30, 100]\\
    dt_{bins} &= [1,2], [2, 3], [3, 5], [5, 10], [10, 20], [20, 30], [30, 60], [60, 100]\\
\end{align*}
where a "bin" represents a range of values. Note that we consider all $(dm, dt)$ bins creating a 2D grid of bins. Then all pairs of two points on the time series were considered, resulting in a  corresponding $(dm, dt)$ pair ($dm$ is difference in magnitude, $dt$ is difference in time). Compiling these $dm$, $dt$ values for each pair of points on the time series, we can fill in a 2D histogram based on the number of values that satisfy that $(dm, dt)$ pair.

\begin{figure}[H]
    \centering
    \begin{subfigure}{.33\textwidth}
    \includegraphics[width=0.8\textwidth]{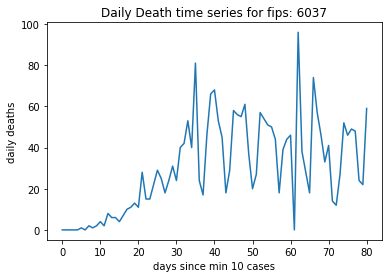}
    \caption{}
    \label{fig:model_6037}
    \end{subfigure}%
    \begin{subfigure}{.33\textwidth}
    \includegraphics[width=0.8\textwidth]{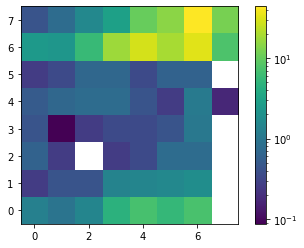}
    \caption{}
    \label{fig:model2_6037}
    \end{subfigure}
    \begin{subfigure}{0.33 \textwidth}
    \includegraphics[width=0.8\textwidth]{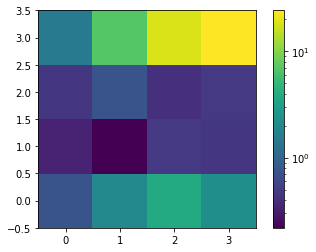}
    \caption{}
    \label{fig:pooled_model_6037}
    \end{subfigure}
    \caption{(Left): Example daily deaths time series (FIPS 6037). (Middle): ($dm$, $dt$) histogram for FIPS county code 6037 for the daily deaths time series since cumulative case count of 10. Note that a LogNorm is applied to the data for plotting purposes , so that all values are between $[0, 1]$ with a log scale (to assign colors to cells). (Right): Average pooled ($dm$, $dt$) histogram for fips county code 6037}
\end{figure}

This time series is turned into a $8 \times 8$ matrix, which when plotted looks like figure \ref{fig:model2_6037}. Each such ($dm$, $dt$) image is then pooled using a (2, 2) block to take an average across the grid (so that the (8, 8) representation is turned into a (4, 4) representation), see figure \ref{fig:pooled_model_6037} for reference. These (4, 4) representations of the daily death time series are flattened into a 16-sized array. These feature representations are given as input to K-Means clustering (with number of clusters set to $6$), and the resulting cluster labels for each county is used as features in various models.

\bigskip

\section{Results}
\label{sec:results}

\subsection{Model Performance}
\label{sec:modelperformance}

All models are evaluated using pinball loss and RMSE (based on the predictions for the 50th quantile) as shown in Table 1. We show results for a 14-day forecast for two periods: Period 1 (May 10, 2020 to May 23, 2020) and Period 2 (May 25, 2020 to June 4, 2020). The relative rankings of models are more or less consistent as we increase forecast-time period, with the LSTM and the ensemble as our best models.

\begin{table}[h!!!!]
\centering
\begin{tabular}{|c|c|c|c|c|}
\hline
\multirow{2}{*}{\textbf{Model}}          & \multicolumn{2}{c|}{\textbf{Period 1}} & \multicolumn{2}{c|}{\textbf{Period 2}} \\ \cline{2-5} 
                                         & \textbf{Pinball Loss}   & \textbf{RMSE}    & \textbf{Pinball Loss}   & \textbf{RMSE}    \\ \hline
Ensemble Model \ref{sec:ensemble}                          & \textbf{0.1054}         & \textbf{1.6391}  & \textbf{0.0896}         & 1.4543  \\ \hline
Gaussian Processes  \ref{sec:gaussianprocess}                     & 0.1560                  & 4.6376           & 0.1038                  & 1.9252           \\ \hline
Imperial Model   \ref{sec:imperialcollegemodel}                        & 0.1337                  & 2.0439           & 0.1012                  & 1.6285           \\ \hline
LSTM              \ref{sec:lstmmodel}                       & 0.1182                  & 1.7416           & \textbf{0.0907}         & \textbf{1.4036}  \\ \hline
Standard Neural Network  \ref{sec:standardneuralnetwork}                & 0.1308                  & \textbf{1.5577}  & 0.0994                  & 1.5068           \\ \hline
Quantile Gradient Boosted Decision Trees \ref{sec:quantilegbdt} & 0.1356                  & 3.2682           & 0.1155                  & 1.9947           \\ \hline
Quantile Neural Network   \ref{sec:quantilenn}               & \textbf{0.1163}         & 1.6557           & 0.0929                  & 1.4639           \\ \hline
Random Forest           \ref{sec:randomforestmodels}                 & 0.1431                  & 3.0462           & 0.0937                  & \textbf{1.4425}  \\ \hline
Random Forest (Moving)                  & 0.1199                  & 1.6663           & 0.1183                  & 1.4834           \\ \hline
SEIR-QD                  \ref{sec:seirmodel}                & 0.1455                  & 1.9874           & 0.1189                  & 1.746            \\ \hline
\end{tabular}
\bigskip
\caption{Model performance measured in pinball loss and RMSE for Period 1 (May 10, 2020 to May 23, 2020) and Period 2 (May 25, 2020 to June 4, 2020). Bolded text indicates lower pinball/rmse score.}
\end{table}

The quantiles predicted by the model on 4 of the counties most affected by the pandemic in Period 1 and Period 2 are shown in Figure \ref{fig:largecountypredictperiod1} and figure \ref{fig:largecountypredictperiod2} below. 10 days of deaths preceding our model  predictions (part of the training set) are also shown on the same plots.

\begin{figure}[h!]
\centering
\begin{tabular}{cc}
\includegraphics[width=0.45\textwidth]{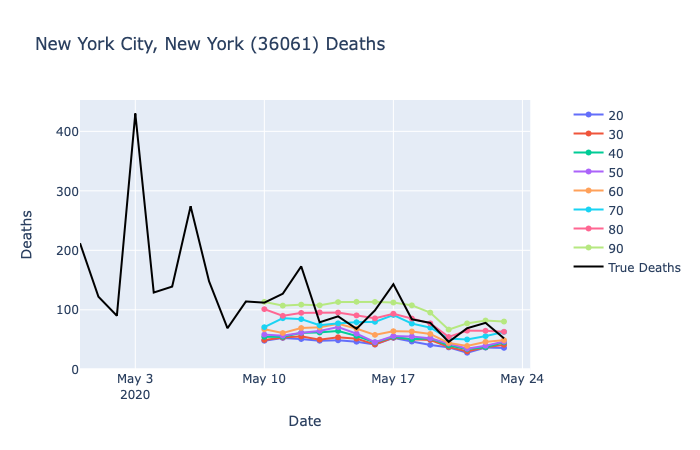} &
\includegraphics[width=0.45\textwidth]{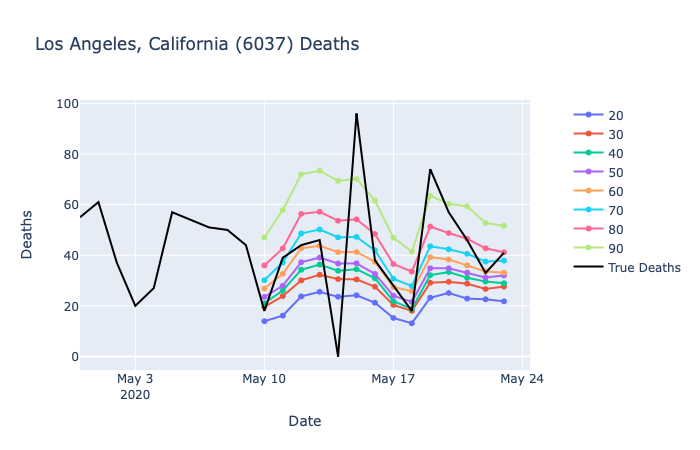} \\
\includegraphics[width=0.45\textwidth]{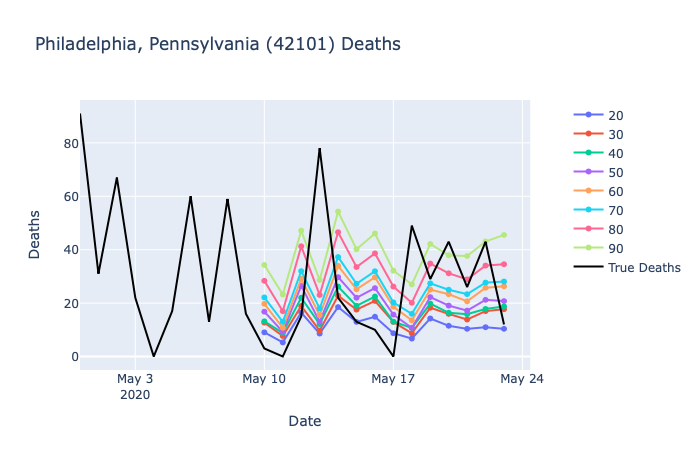} &
\includegraphics[width=0.45\textwidth]{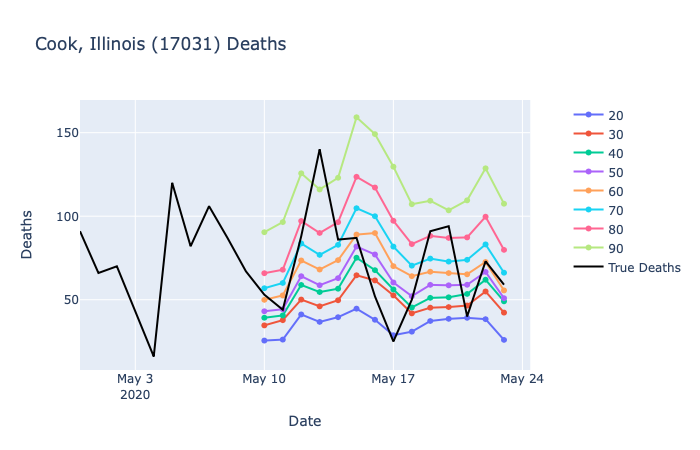} \\
\end{tabular}
\caption{Period 1 ensemble model predictions on FIPS county codes: 36061, 6037, 42101, 17031}
\label{fig:largecountypredictperiod1}
\end{figure}

\begin{figure}[h!]
\centering
\begin{tabular}{cc}
\includegraphics[width=0.45\textwidth]{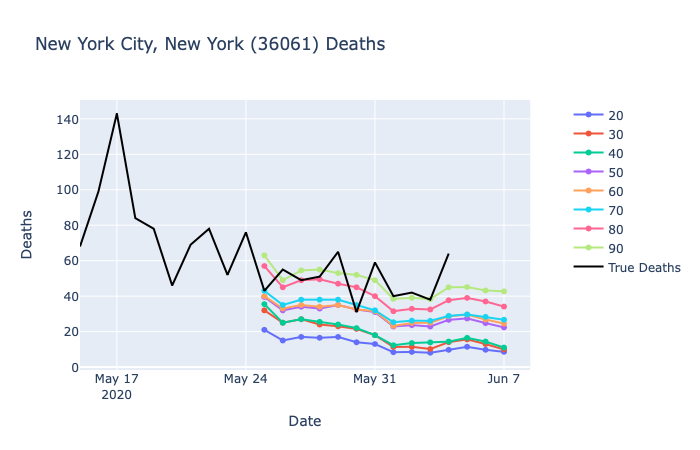} &
\includegraphics[width=0.45\textwidth]{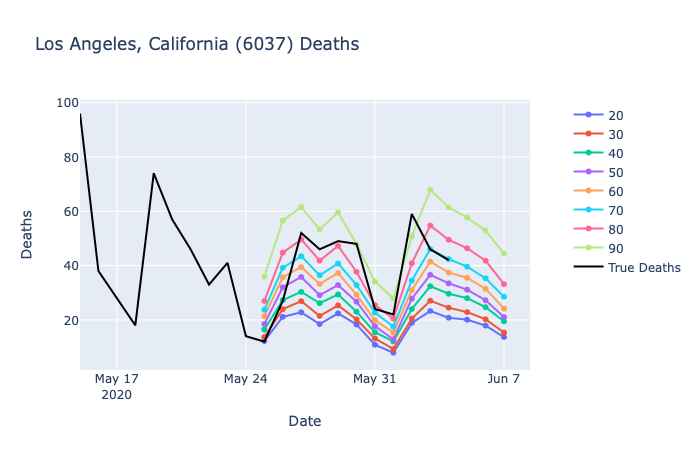} \\
\includegraphics[width=0.45\textwidth]{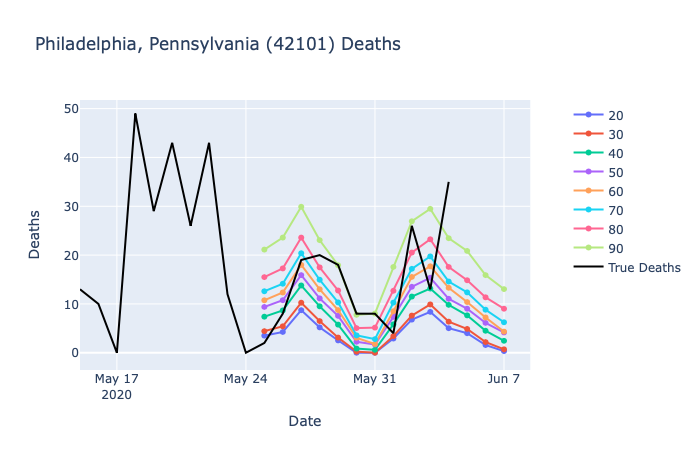} &
\includegraphics[width=0.45\textwidth]{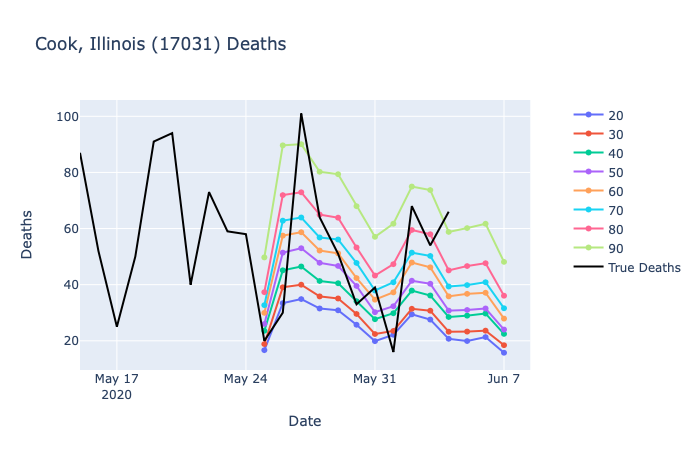} \\
\end{tabular}
\caption{Period 2 ensemble model predictions on FIPS county codes: 36061, 6037, 42101, 17031}
\label{fig:largecountypredictperiod2}
\end{figure}

\newpage

\section{Discussion and Conclusion}
\label{sec:discussion}

As outlined in the paper, the best performing model is ultimately an ensemble of epidemiological and machine learning models. Our initial focus was on epidemiological techniques, the predominant class of models used in epidemiology to model pandemics . While these models are able to generate stable prediction distributions over longer time horizons, they are unable to take full advantage of all of the available data sources to predict finer resolution trends in the death data.

As the pandemic progressed, our models got to train on more data. These factors motivated us to explore machine learning models that would require minimal injection of priors on the dynamics of the disease spread, and instead rely on learning patterns in the data. Sequence models (described in \ref{sec:lstmmodel}) show the best performance as expected, but even non-time-series models such as tree-based models (\ref{sec:quantilegbdt}, \ref{sec:randomforestmodels}) and standard neural networks (\ref{sec:standardneuralnetwork}, \ref{sec:quantilenn}) are able to capture daily fluctuations in the data, while maintaining reasonable predictions over longer horizons.

While most machine learning models are difficult to interpret, some of them offer information on the importance of the various input features in predicting death rates. Lagged dynamic features of cases, deaths and mobility are the most important variables, with lagged cases being the most important among them. In addition, temporal features such as date offsets, and day of the week are important, presumably helping with differentiating which stage of the pandemic a region is in (as well as capturing weekly fluctuations in reporting, or the so-called "Sunday effect"). Static features are much less important, but certain density features such as population density had some predictive power.

To the best of our knowledge, this is the first geographically local (county level), variable-length forecast, US COVID-19 daily death model combining epidemiological and machine learning methods. We have found a model that provides a more complete description of the US COVID-19 daily death distribution, and has a consistently low RMSE score on daily deaths across different time periods. We hope this work can be useful for local/state level policymakers, as they decide how to deal with the ongoing pandemic in the US, and how to prepare for future waves of COVID-19-like epidemics.

\end{document}